\documentclass[letterpaper, 10 pt, conference]{ieeeconf}  %

\IEEEoverridecommandlockouts                              %

\usepackage{graphicx} %
\usepackage{times} %
\usepackage{amsmath} %
\usepackage{amssymb}  %
\usepackage{siunitx}  %

\title{\LARGE \bf
Risk-Aware Navigation for Mobile Robots in Unknown 3D Environments}

\author{Elie Randriamiarintsoa $^{1}$, Johann Laconte $^{2}$, Benoit Thuilot$^{1}$, Romuald Aufrère$^{1}$ %
\thanks{$^{1}$Université Clermont Auvergne, Clermont Auvergne INP, CNRS, Institut Pascal, F-63000 Clermont-Ferrand, France.
{\tt\small elie.randriamiarintsoa@uca.fr}}%
\thanks{$^{2}$University of Toronto Institute for Aerospace Studies (UTIAS), 4925 Dufferin St, Ontario, Canada.
{\tt\small johann.laconte@utoronto.ca}}%
}
\usepackage{xcolor}
\usepackage{lipsum}

\usepackage{csquotes}
\usepackage{hyperref}
\usepackage{pdfpages}
\usepackage[english]{babel}

\usepackage{amsfonts}
\usepackage[
backend=biber, 
bibencoding=utf8,
style=ieee, 
sorting=none, 
natbib=true, 
doi=false, 
isbn=false, 
url=false, 
eprint=false, 
maxcitenames=1, 
mincitenames=1,
]{biblatex}

\usepackage{acronym}
\acrodef{DEM}{Digital Elevation Map}
\acrodef{BOF}{Bayesian Occupancy Filter}
\acrodef{NMPC}{Nonlinear Model Predictive Control}
\acrodef{UAV}{Unmanned Aerial Vehicle}
\acrodef{CVaR}{Conditional value-at-risk}
\acrodef{AIS}{Automated Information System}

\newcommand{\Path}{\mathcal{P}}

\begin{document}

\maketitle
\thispagestyle{empty}
\pagestyle{empty}

\begin{abstract}
    Autonomous navigation in unknown 3D environments is a key issue for intelligent transportation, while still being an open problem. %
Conventionally, navigation risk has been focused on mitigating collisions with obstacles, neglecting the varying degrees of harm that collisions can cause.
In this context, we propose a new risk-aware navigation framework, whose purpose is to directly handle interactions with the environment, including those involving minor collisions.
We introduce a physically interpretable risk function that quantifies the maximum potential energy that the robot wheels absorb as a result of a collision.
By considering this physical risk in navigation, our approach significantly broadens the spectrum of situations that the robot can undertake, such as speed bumps or small road curbs.
Using this framework, we are able to plan safe trajectories that not only ensure safety but also actively address the risks arising from interactions with the environment.
\end{abstract}

\section{INTRODUCTION}\label{sec:intro}
\begin{figure}[t]
  \centering
  \includegraphics[width=\linewidth]{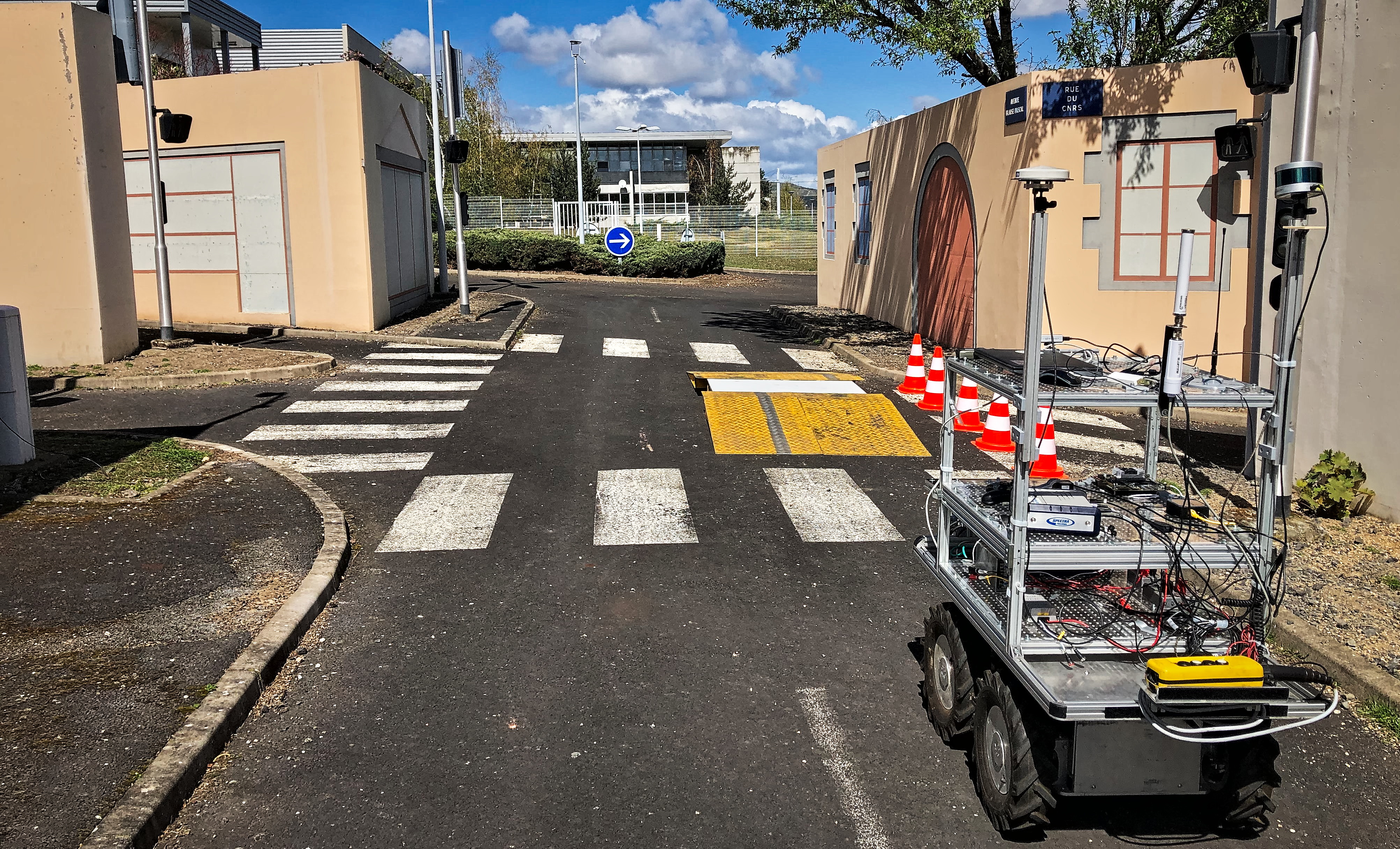}
  \caption{Example of a situation that a vehicle might encounter while navigating.
    Speed bumps are frequent obstacles an intelligent vehicle has to safely overcome.
    With our framework, the vehicle is able to make more effective decisions based on the level of risk it is allowed to take.
}
  \label{fig:intro}
  \vskip-.5em
\end{figure}

    Navigating in unknown 3D environments is a crucial task for mobile robots.
Before being able to move, a robot must represent the environment in which it evolves.
A well-known and efficient way to map the environment is the occupancy grids introduced by \citet{elfes1989using}.
An occupancy grid provides the robot with information about the potential presence of an obstacle at a given position.
However, this information alone does not encompass all the capabilities of the robot to maneuver within its environment. 
For example, a wheeled robot can safely cross grasses, a curb, or a speed bump at low speed, as shown in \autoref{fig:intro}.
Therefore, a way to assess their hazardous nature is necessary to evolve in these environments.
Motivated by this fact, numerous recent works \cite{shi2023gridcentric} have contributed to risk-aware navigation.
However, \citet{laconte2019lambda} demonstrated that the Bayesian occupancy grid, which stores the probability of collision of a given position, is ill-suited to compute the risk over paths.
Indeed, the probability of collision, computed as the joint probability that every cell is free of obstacles, is highly dependent on the grid tessellation size.
To overcome this difficulty, they developed a novel framework called Lambda-Field to assess physics-based risks on occupancy grids.
Moreover, in occupancy grids framework, most navigation methods \cite{PATLE2019582} use geometric and semantic reasoning to handle the obstacles present in the environment, such as curbs, traffic cones, speed bump, sidewalk, buildings, traffic circle and traffic lights shown in \autoref{fig:intro}.
\citet{laconte2019lambda} showed that path planning becomes more intuitive and meaningful when a physical risk metric is used.
However, their work has been developed solely for 2D environments.
Here, we adapt their work to 3D environments and use physical, risk-based reasoning to deal with the obstacles represented in \autoref{fig:intro}.

The main contributions of this paper are i) an extension of the Lambda-Field \cite{laconte2019lambda} to 3D environments; ii) a generalization of the risk that take into account traversable obstacles; iii) a mathematical formulation of a local risk-aware path planning algorithm in the Lambda-Field; iv) a demonstration of the applicability of the method, using data acquired in real urban environment.

\section{RELATED WORK}\label{sec:related_work}
    An important challenge in local motion planning is to find a feasible path that matches the robot's capabilities while being safe to traverse.
Prior to identifying such a path, it is essential to accurately map the immediate surroundings of the robot, thereby representing its local environment.
A popular representation of the robot surroundings is the occupancy grids, introduced by \citet{elfes1989using}.
The main idea of the occupancy grids is to tessellate the environment into cells and to store in each cell the information of occupancy.
\citet{coue2006bayesian} enhanced the previous idea by adding a Bayesian layer to the approach.
Their work led to the \ac{BOF} techniques.

Using these Bayesian occupancy grids, standard motion planning frameworks plan collision-free paths by assessing the probability of collision of each potential robot path.
However, assessing only the risk of colliding with an obstacle will always lead the robot to go around them.
Therefore, the collision risk does not fully exploit the robot ability.
Indeed, a mobile robot can cross tall grass or a speed bump even if it results in collisions.
To allow robots to make more insightful choices, several approaches propose to integrate the notion of risk in motion planning by using risk maps.
Unlike conventional Bayesian occupancy grids, where each cell stores the occupancy probability of a given position, a risk map stores the risk at this position.
\citet{schroder2008path} designed a risk map for cognitive vehicles.
The defined risk map is used to prevent the robot from approaching the hazardous obstacles, such as pedestrians or cars.
\citet{pereira2011toward} used historical shipping traffic and bathymetry data of coastal regions to create a risk map for underwater vehicles.
Assigning a probabilistic risk value to each position that is likely to be occupied allows to avoid potential hazardous collisions.
\citet{primatesta2019risk} used a risk map to quantify the risk of an unmanned aerial vehicle flying over a given position to cause lethal incidents in an inhabited area.
Even though these techniques produce good results, they all make the assumption that the risk is only dependent on the robot position, while intuitively the risk depends also on the robot state and capabilities.
As such, our work proposes a risk model that exploits both the state of the environment and the robot ability to assess the risk of a given path.

In risk-aware navigation, \citet{majumdar2020should} provide insight about which metrics should be used.
Among them, the expected value and the \ac{CVaR} are identified as valid candidates.
The \ac{CVaR} is used to capture the worst-case expected hazardous event that could happen over a given horizon of time, and has been used by several recent works \cite{rockafellar2000optimization,ROCKAFELLAR20021443}.
\citet{hakobyan2019risk} measure the risk of colliding with randomly moving obstacles.
They define the problem of path planning as a constrained optimization where the distance to the obstacles must be superior or equal to a fixed threshold.
They demonstrate that this formulation allows to plan an effective path for a quadrotor in a 3D environment, while adjusting the safety and prudence of the motions.
For mobile robots, \citet{fan2021step} enhance the previous work by adding multiple sources of risk, such as the slippage risk.
A new risk map is then created by aggregating the risks that come from these different sources.
\citet{cai2022risk} quantify the risk by converting a learned speed distribution map to a risk cost value.
In constrast to our work, traversability is captured via experienced trajectories.
\citet{koval2022experimental} propose another risk-aware path planning which is also based on a priori data, precisely on a known map.
In \cite{laconte2019lambda}, the authors introduce the Lambda-Field, a generic method to assess a physical risk over a continuous path.
The risk is defined as the expected force of collision along a path.
They introduced a new mapping framework where each cell stores the density of collisions that could be hazardous to the robot.
In contrast to the Bayesian occupancy grids framework, the Lambda-Field framework computes the integral of the risk over a given path.
This approach has the ability to retain the physical units of the risk.
They have demonstrated the relevance of the approach in 2D static environments.
They extended their work to unstructured \cite{Laconte_2021} and dynamic environments \cite{laconte2021dynamic}, yet still only considering 2D obstacles.
In this article, we adapt their framework to 3D static environments and propose a new physically coherent risk measure.

\section{PRELIMINARIES}\label{sec:preliminaries}
    In this section, we briefly present the framework developed by \citet{laconte2019lambda}.
In a similar way as occupancy grid frameworks, the environment is tessellated into cells, where all cells have a fixed size and area $\Delta a \in \mathbb{R}_{>0}$.
The core difference of their approach, compared to a standard Bayesian occupancy grid, is the information that is stored in each cell.
Instead of estimating the probability of collision, they estimate the density of collision $\lambda \in \mathbb{R}_{\geq 0}$ for each cell, also called the intensity.
The probability of collision within the cell is then $\lambda \Delta a$. The higher the intensity of a cell, the more likely it is that a hazardous event will happen in this cell.
On the contrary, an intensity of zero means that the cell will never lead to a harmful collision and can be crossed safely.
As shown in \cite{laconte2019lambda}, the probability of facing at least one hazardous event (i.e., collision) over a path $\mathcal{P}$ in the discretized field is
\begin{equation}
    \label{eq:P_coll_Discrete}
    \mathbb{P}(\text{coll}|\mathcal{P}) = 1 - \exp\left(-\Delta a \sum_{c_{i} \in \mathcal{C}}\lambda_i\right),
\end{equation}
where $\mathcal{C}$ is the set of cells crossed by the path $\mathcal{P}$, and $\lambda_i$ is the intensity of a cell $c_i \in \mathcal{C}$.

In order to build this map, \citet{Laconte_2021} use a 2D lidar.
If $e$ is the error region area attached to every lidar measurement, then a cell $c_i$ is measured as hazardous if it is located within the error region area of a lidar beam impact.
Otherwise, if the lidar beam traversed the cell without returning a collision, the cell is measured as safe.
Under these considerations, the intensity $\lambda_i$ of each cell is computed using an expectation maximization approach. %
Let $s_i$ be the number of times a cell $c_i$ is measured as safe and $h_i$ the number of times a cell $c_i$ is measured as hazardous, it can be shown, see \cite{Laconte_2021}, that the intensity of the cell $c_i$ can be expressed as
\begin{equation}
    \lambda_i = \frac{1}{e}\ln\left(1+\frac{h_i}{s_i}\right).
    \label{eq:measur_2D}
\end{equation}

Furthermore, the main advantage of the Lambda-Field is that it enables the computation of the probability of collision, but also of a generic risk over a path.
Following the demonstration in \cite{Laconte_2021}, the expected value of a generic deterministic risk over a path $\mathcal{P}$ crossing the cells $\{c_i\}_{0:N-1}$, is given by
\begin{equation}
    \mathbb{E}[r(X)] = \sum_{i=0}^{N-1} K_i r(x_i),
    \label{eq:E_risk}
\end{equation}
where $x_i$ corresponds to the position on the path associated to the cell $i$ and $r(x_i)$ is a generic risk function that gives the value of the deterministic risk if a collision would happen at $x_i$.
$K_i$ gives the probability of encountering an harmful event at $x_i$ and is given by
\begin{equation}
    \label{eq:Ki}
    K_i = \exp\left(-\Delta a \sum_{j=0}^{i-1} \lambda_j\right)(1-\exp(-\Delta a \lambda_i)).
\end{equation}
$\mathbb{E}[r(X)]$ encompasses both the state of the environment through the intensity field, and the state of the robot when it comes at a given position along the path, though the risk function $r(x_i)$.
As such, this framework provides a way to compute a meaningful risk over a path in occupancy grids.

\section{MAPPING}\label{sec:mapping}
We present here our extension of the Lambda-Field to take into account traversable 3D obstacles in the environment.
For that, we use a 3D lidar to compute a \ac{DEM} \cite{DEM}.
By aggregating multiple point clouds over time, we compute the difference in elevation of the environment.
This is achieved by taking into account the elevations of the eight neighboring cells for each grid cell.
Then, this \ac{DEM} is used to compute the Lambda-Field.

To model traversable obstacles in the context of urban environments containing road curbs and speed bumps.
We define a cell $c_i$ as safe if the value of the difference in elevation is below a threshold value $H_\text{safe}$; otherwise, this cell is defined as hazardous.
The intensity equation \autoref{eq:measur_2D} is however modified to take into account the severity of the event and will stop the robot in its course.
As such, the intensity of a cell $c_i$ is computed with
\begin{equation}
    \lambda_i = \frac{1}{e}\ln{\left(1+\frac{h_i}{s_i}\right)}p_{i}
    \label{eq:3D_lambda_field}
\end{equation}
where quantity $p_i$ describing the severity of this collision is defined by
\begin{equation}
    \label{eq:harmful_prob}
    p_{i} = \min\left(\frac{|H_i|}{R}, 1\right),
\end{equation}
with $H_i$ the difference in elevation of the cell $c_i$ with reference to its neighbors and $R$ the radius of the wheel.
As such, the likelihood it stop $p_i$ is one when the obstacle is higher than the wheel radius and the robot has no chance to go over the obstacle, and tends to zero for small obstacles which are harmful.
In the conservative case where we assume the robot is not able to go over any obstacle, then $p_i=1$ and the measurement equation returns to \autoref{eq:measur_2D}.
\autoref{eq:3D_lambda_field} is then a generalization of the approach where we consider that the robot is also able to traverse over small obstacles.

\begin{figure}[t]
  \centering
  \includegraphics[width=\linewidth]{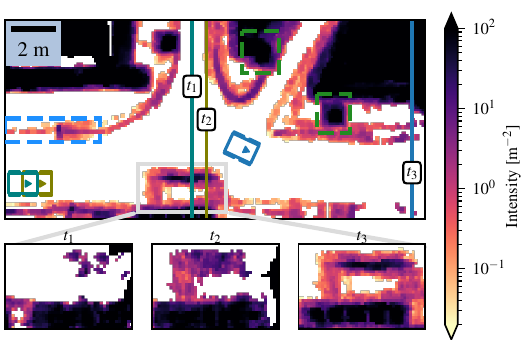}
  \caption{Example of Lambda-Field of the environment depicted in \autoref{fig:intro}.
  The area containing the speed bump and the cones is enlarged and shown at different times.
  The lidar range (vertical line) and the robot (box) pose for each of these times are illustrated in teal ($t_1$), dull green ($t_2$) and blue ($t_3$).
  The left curb (light blue) and the traffic lights (green) seen in \autoref{fig:intro} are outlined in dashed lines.}
  \label{fig:lambda-field_3D}
  \vskip-.5em
\end{figure}

As an example, \autoref{fig:lambda-field_3D} depicts a Lambda-Field computed using \autoref{eq:3D_lambda_field}, for the environment shown in \autoref{fig:intro}.
The global Lambda-Field map is shown at the top in \autoref{fig:lambda-field_3D} and the construction of the map around the speed bump (in yellow in \autoref{fig:intro}) at different times is illustrated below.
At time $t_1$, the speed bump is in the lidar range and the intensities $\lambda_i$ representing the speed bump began to converge.
The speed bump is better represented at time $t_2$ through new and additional measurements.
As the speed bump has been completely scanned at time $t_3$, the intensities associated with it have completely converged.
One can note that the curbs, traffic cones, speed bump, buildings and traffic lights poles are well detected.
The borders of the speed bump have high intensity values, when its ascending and descending portions have more nuanced intensities, meaning that harmful collisions are less prone to happen.
As a result, the robot will be able to cross the speed bump by taking a reasonable risk.
Additionally, buildings, traffic light poles, and traffic cones have high intensities.
These obstacles cannot be crossed by the robot without taking an unreasonable risk.

\section{RISK ASSESSMENT}\label{sec:risk}
    In order to navigate in the Lambda-Field, we define a risk function that reflects the potential obstacles that the robot can face. We choose to describe the risk by the maximum potential energy absorbed by the wheels.

\begin{figure}[!htbp]
  \centering
  \includegraphics[width=\linewidth]{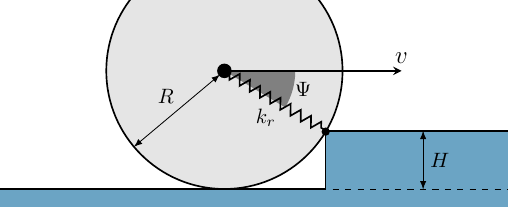}
  \caption{Modeling of the wheel of radius $R$ in collision with the curb (in blue) of height $H$, with a speed $v$ and angle $\Psi$.
  The deformation of the wheel due to the collision is approximated with the deformation of a spring of stiffness $k_r$.}
  \label{fig:elevation_risk}
  \vskip-.5em
\end{figure}

We illustrate in \autoref{fig:elevation_risk} the modeling of the wheel.
The wheel is modeled as a deformable disk, see \autoref{fig:elevation_risk}.
When the wheel crosses an elevated obstacle, it deforms at the point where the two objects collide.
This deformation is approximated with the deformation of a spring of stiffness $k_r$.
In order to find the maximum amplitude $l_{m}$ of the tire compression due to the collision, we solve the following differential equation:
\begin{equation}
    \label{eq:eq_diff_spring}
    \ddot{l} + \omega^2l = 0 \quad \text{with } \omega = \sqrt{\frac{k_r}{m}}
\end{equation}
where $l$ is the compression at the contact point and $m$ is the mass of the vehicle.
As such, the maximum amplitude $l_{m}$ of the tire compression is
\begin{equation}
    \label{eq:max_defflection_spring}
    l_{m} = \frac{v\cos(\Psi)}{\omega}
\end{equation}
where $v$ is the linear velocity of the robot and ${\Psi=\arcsin(\frac{R-\min(H,R)}{R})}$ is the angle of attack of the collision.
The angle $\Psi$ depends on the elevation of the obstacle and the radius $R$ of the wheel.
To be conservative, we take the maximum difference in elevation $H$ of the obstacles that lie in the transverse axis of the robot path $\mathcal{P}$.
Finally, the risk function is defined as the maximum potential energy that will be absorbed in the wheels at the time of collision and is computed by
\begin{equation}
    \label{eq:risk_function}
    r(X) = \dfrac{1}{2} k_r  l_{m}^2
\end{equation}
As such, the risk function $r(X)$ has a clear physical meaning and is expressed in Joule.
In the next section, we use this risk assessment to construct safe paths in 3D environments.
    
\section{PATH PLANNING}\label{sec:planning}
    The goal of the robot is to follow a local reference path $\Path_{ref} \subset \mathbb{R}^3$ that is assumed to be known.
However, unforeseen events, detected by the robot's sensors, can make this path impassable.
The aim of our path planning framework is to find the optimal path $\Path^* \subset \mathbb{R}^3$ that is the closest to the reference path $\Path_{ref}$, while maintaining a reasonable level of risk along the path.
Namely, a path $\Path$ is accepted only if the risk undertaken by the robot along the path is below a user-defined risk threshold $r_{threshold}$, consistent with the preferences of the user and the robot abilities.
The optimal path $\Path^*$ also minimizes the traversal time of the robot.

To address the problem, we rely on a \ac{NMPC} approach, where we use the Ackermann steering model as the prediction model.
Using a small sample time $\Delta t$, the kinematic equations can be discretized into
\begin{equation}
	\begin{aligned}
		\begin{bmatrix}
			x_{k+1} \\y_{k+1} \\ \theta_{k+1}
		\end{bmatrix}
		=
        \begin{bmatrix}
            x_{k}\\
			y_{k}\\
			\theta_k
        \end{bmatrix}
        + \Delta t
		\begin{bmatrix}
			v_{k} \cos{\theta_{k}}\\
			v_{k} \sin{\theta_{k}}\\
			v_{k} \tan{\delta_{k}}/L
		\end{bmatrix}
        ,
	\end{aligned}
	\label{eq:discrete_model}
\end{equation}
where $k$ represent the discrete time, $L$ is the length of the wheelbase of the robot, $(x_{k}, y_{k})$  and $\theta_k$ are respectively the centre position of the rear axle and the yaw angle of the robot in the absolute frame.
Finally, $v_k$ is the linear velocity of the robot, and $\delta_k$ the steering angle of the robot.
The compact form of \autoref{eq:discrete_model} is written as $\textbf{x}_{k+1} = f_d(\textbf{x}_k, \textbf{u}_k)$, where $\textbf{x}_{k+1} = [x_{k+1}, y_{k+1}, \theta_{k+1}]^\intercal$ is the predicted state vector, $\textbf{x}_k = [x_k, y_k, \theta_k]^\intercal$ is the state vector, and $\textbf{u}_k = [v_k, \delta_k]^\intercal$ is the control vector.
The reference trajectory associated with the path $\Path_{ref}$ is defined as a set of desired states $[\textbf{x}^{r}_{0}, \cdots, \textbf{x}^{r}_{N_{p}}]^\intercal$, where $\textbf{x}^{r}_{k}$ is the desired state at time $k$.%

According to our objective and constraints, we want to minimize the errors between the predicted and the reference paths over a finite prediction horizon of length $N_{p}$ while considering kinematic, control, and risk constraints.
We also want to minimize the traversal time and penalize the error in the final state.
The penalization of this latter error is used to force the robot to approach the farthest reference goal $\textbf{x}^{r}_{N_{p}}$ located in the perceived and therefore secure navigable area.
As a result, the risk-aware navigation based on \ac{NMPC} consists in computing a trajectory $[\textbf{x}_{0}, \cdots, \textbf{x}_{N_{p}}]^\intercal$ by the following quadratic constraint optimization:
\begin{equation}
    \begin{aligned} 
        \min_{} \quad &Z = Z_{1} + Z_{2} + Z_{3}\\
         \textrm{s.
t.
} \quad & \textbf{x}_{k+1} = f_d(\textbf{x}_k, \textbf{u}_k)\\
           &\textbf{u}_{\rm min} \le \textbf{u}_k \le \textbf{u}_{\rm max}\\
           &\mathbb{E}[\textit{r}(X)] \leq r_{threshold}
    \end{aligned}
 \label{eq:prob_control}
 \end{equation}
where $Z$ is the cost function of the optimization problem, defined as a combination of three penalty terms described below, and $(\textbf{u}_{min}, \textbf{u}_{max})$ are respectively the lower and upper bounds of the control vector.

The first penalty term is used to prevent the robot from deviating too far from the reference path $\Path_{ref}$:
\begin{equation}
    \label{eq:penalty_track}
    Z_{1} = \sum_{k=0}^{N_{p}-1} (\textbf{x}_k-\textbf{x}_k^r)^\intercal \textbf{Q} (\textbf{x}_k-\textbf{x}_k^r),
\end{equation}
where $\textbf{Q} \in \mathbb{R}^{3\times3}$ is the weight matrix used to penalize the state component parts.
The second penalty term ensures that the robot approaches to the farthest desired position:
\begin{equation}
    \label{eq:penalty_goal}
    Z_{2} = (\textbf{x}_{N_{p}}-\textbf{x}_{N_{p}}^r)^\intercal \textbf{Q}_{N} (\textbf{x}_{N_{p}}-\textbf{x}_{N_{p}}^r),
\end{equation}
where $\textbf{x}_{N_{p}}$ is the last predicted state, and $\textbf{Q}_{N} \in \mathbb{R}^{3\times3}$ is the weight matrix used to penalize the final state component parts.
The last penalty term is used to prioritize a minimum traversal time:
\begin{equation}
    \label{eq:penalty_time}
    Z_{3} = \sum_{k=0}^{N_{p}-1} \text{w}_{v}(v_k - v_{max})^2,
\end{equation}
where $v_{max}$ is the maximum velocity the robot can reach and $\text{w}_{v} \in \mathbb{R}$ is the weighting coefficient that penalizes low velocities.
The final constraint $\mathbb{E}[\textit{r}(X)] \leq r_{threshold}$ forces the robot to find a trajectory that is safe to traverse.
Therefore, any trajectory that exceeds the risk threshold $r_{threshold}$ will be rejected.
Note that we do not add the risk to the cost function, as we would lose its physical meaning.
The risk is here defined as a hard constraint to ensure that it is never exceeded.
Finally, the optimal path $\Path^*$ is the one that minimizes the cost function in respects to the risk and the control constraints.

\section{VALIDATION OF THE FRAMEWORK}\label{sec:exp}
    \begin{figure*}[!htbp]
      \centering
      \includegraphics[width=\linewidth, ]{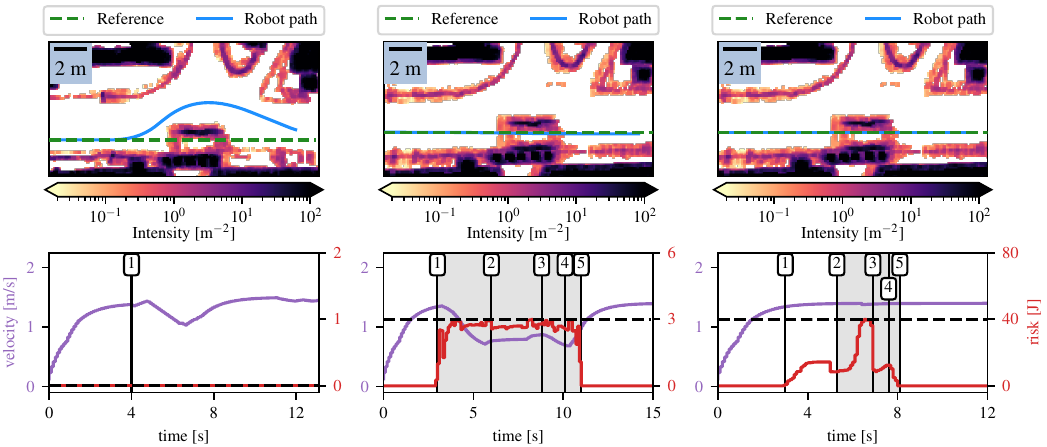}
      \caption{Example of an environment where an unexpected obstacle (speed bump) occurred on the reference path (dashed green).
      We investigated three risk thresholds, \SI{0}{\J} (left), \SI{3}{\J} (middle) and \SI{40}{\J} (right).
      For each threshold, the path taken by the robot is in blue.
      Its velocity and risk are depicted in purple and red.
      Events:
      (\textbf{1}): Speed bump came into the robot's perception field.
      (\textbf{2}): Robot started to climb the speed bump.
      (\textbf{3}): Robot reached the speed bump top.
      (\textbf{4}): Robot started to get down.
      (\textbf{5}): Robot reached the asphalt.
      The gray shaded areas show the duration when the robot traversed the speed bump.
  }
      \label{fig:simu}
      \vskip-.5em
\end{figure*}
To show the applicability of this framework, we ran three scenarios in the environment depicted in \autoref{fig:intro}, in which three different risk thresholds were considered.
We used real data for the perception part and simulated the path planning with a simulation model of the robot seen in \autoref{fig:intro}.
The simulations were performed on a computer equipped with an 11th Gen Intel Core i7-11850H processor and an NVIDIA GeForce RTX 3080 graphics card.
In all scenarios, the robot starts in front of the speed bump with zero steering angle and zero speed.
The Lambda-Field construction involves setting the difference in elevation threshold $H_\text{safe}$ to \SI{5}{\cm} and the $e$ value to $\SI{1}{\cm\squared}$.
Each Lambda-Field grid map consist of $200\times200$ cells, where each cell has a size of $10\times\SI{10}{\cm}$, resulting in a map of $20\times\SI{20}{\m}$.
The simulated robot has a wheel radius $R$ of \SI{25}{\cm} and a mass $m$ of $\SI{50}{\kg}$. The stiffness $k_r$ of each robot wheel is set to \SI{150000}{\N/\m}.
The maximum velocity $v_\text{max}$ of the robot is set to \SI{1.5}{\m/\s}, while the steering angle range $(\delta_{\text{min}},\delta_\text{max})$ is set to $\pm\SI{11}{\degree}$.
The sampling time $\Delta t$ is set to \SI{100}{\ms}.
The weight matrix $\textbf{Q}$ is a diagonal matrix of elements $[0.05, 0.05, 0.05]$.
The weight matrix $\textbf{Q}_N$ is a diagonal matrix of elements $[1, 1, 1]$.
Therefore, if an unexpected obstacle prevents the robot from tracking the reference path, but the reference goal can be reached by a path deviating from the reference path, then the path planner will be able to choose this path.
Finally, the weight coefficient $\text{w}_{v}$ is equal to $0.1$.

We show the results in \autoref{fig:simu}, where each column corresponds to the result of a scenario.
The first row in \autoref{fig:simu} shows the path of the robot as well as the reference path, which is a straight line passing through the speed bump.
The second row in \autoref{fig:simu} shows the velocity profile and the risk during the traversal.
The graphs are labeled with numbers corresponding to different times of interests:
    $t_1$: the speed bump is detected;
    $t_2$: the robot ascend the speed bump;
    $t_3$: the robot reaches the top speed bump;
    $t_4$: the robot descends the speed bump; and
    $t_5$: the robot comes back to the road.

In the first scenario, the risk threshold is set to zero, meaning that the robot is not allowed to take any risk.
As long as the robot does not see the speed bump, the reference path is tracked.
At time $t_1$, the robot can no longer track the reference path, as crossing the speed bump is risky.
The path planning formulation leads the robot to go around the speed bump.
One can note that throughout the traversal, the robot maximizes its velocity without ever crossing a potential risky cell, as no risk is allowed.
This scenario shows that setting the risk to zero is the same as assessing the risk of collision and imposing that the robot goes around the obstacles.

However, in some cases, avoiding may not be possible (for example, the left lane is forbidden due to traffic rules or occupied by another vehicle) and if no risk is allowed the robot would then stop in front of the speed bump.
With our approach, it is possible to reach the goal by managing the risk threshold in accordance with the capability of the robot.
In the scenario shown in the second column in \autoref{fig:simu}, we set the risk threshold $r_{threshold}$ to \SI{3}{\J}, meaning in our case that \SI{6}{\mm} compression of the wheel is the maximum allowed.
One can see that the robot passes through the speed bump, but at a low speed.
The robot slows down at time $t_1$, as the speed bump is detected by the robot.
Then, from time $t_2$ to time $t_5$, our path planner regulates the speed of the robot to stay below the risk threshold.
The path planner maintains an adequate speed to cross the speed bump safely, finding a balance between minimizing the traversal time and the risks taken by the robot.
From time $t_3$ to $t_4$, the robot is on the speed bump.
Before leaving it, the robot again detects hazardous events on the reference path, leading the robot to slow down.
At time $t_4$, the robot starts to descend the speed bump at a reasonable speed to manage the risk that has appeared on the path.
Finally, at time $t_5$, the robot leaves the obstacle, accelerating again.
This scenario shows that the robot can handle a consistent amount of risk in complicated situations.

In the last scenario, we increase the risk threshold to \SI{40}{\J}, meaning that we accept a \SI{23}{\mm} maximum tire's compression.
Intuitively, the robot goes faster than in the previous scenario.
At time $t_1$, the speed bump is detected. 
The planner stabilizes the velocity of the robot to traverse the speed bump, while not expecting a risk exceeding \SI{40}{\J}.
Until it reaches its target, the robot tracks the reference path at its greatest allowable velocity.
As such, our framework is able to produce meaningful paths by considering the potential hazards associated with navigating through 3D obstacles.

\section{CONCLUSION}\label{sec:conclusion}
    In this paper, we presented a method of risk-aware navigation in unknown 3D environments.
We propose a generalization of the work of \citet{laconte2019lambda}, taking into account the traversal of small obstacles such as road curbs or speed bumps.
We showed how to compute the associated Lambda-Field, using 3D lidar measurements.
The resulting map is used to evaluate the expected maximum potential energy that the robot's wheels will absorb in the event of a collision along a given path.
Finally, we presented a path planning formulation that take the risk into account as a hard constraint.
Using our formulation, we showed that the risk function is well-fitted for risk-aware navigation in urban environment.
While being able to mimic a standard path planning approach by setting the risk threshold to zero, our framework can also generate a path going over obstacles if the risk is tolerable.
This work focused on explaining the theoretical framework and its applicability.

Future works will focus on improving the path planning part and demonstrating the pertinence of the framework in larger-scale experiments.
The exploration of alternative risk metrics, such as \ac{CVaR}, will be pursued to account for tail events.
We intend to extend our framework to enable the robot to perform long-term missions.
Indeed, providing mobile robots with such capabilities will augment the autonomy of intelligent vehicles in rural environments.
A global path planner based on OpenStreetMap \cite{open_street_map} will be included to the framework.
Furthermore, we will conduct extensive experiments on the framework, incorporating quantitative evaluations.
Finally, we will investigate the possibility of adding several risks to further constrain the path-planning algorithm, such as the risk of crossing a continuous lane marking.

\section*{ACKNOWLEDGMENT}
This work has been supported by the AURA Region and the European Union (FEDER) through the LOG SSMI project of CPER 2020 MMaSyF challenge.

\renewcommand*{\bibfont}{\small}
\printbibliography

\end{document}